# APPLICATION OF DEEP LEARNING IN GENERATING DESIRED DESIGN OPTIONS: EXPERIMENTS USING SYNTHETIC TRAINING DATASET

Zohreh Shaghaghian [1], Wei Yan [1]
[1] Texas A&M University, College Station, United States

## ABSTRACT

Most design methods contain a forward framework, asking for primary specifications of a building to generate an output or assess its performance. However, architects urge for specific objectives though uncertain of the proper design parameters. *Deep Learning* (DL) algorithms provide an intelligent workflow in which the system can learn from sequential training experiments. This study applies a method using DL algorithms towards generating demanded design options. In this study, an object recognition problem is investigated to initially predict the label of unseen sample images based on training dataset consisting of different types of synthetic 2D shapes; later, a generative DL algorithm is applied to be trained and generate new shapes for given labels. In the next step, the algorithm is trained to generate a window/wall pattern for desired light/shadow performance based on the spatial daylight autonomy (sDA) metrics. The experiments show promising results both in predicting unseen sample shapes and generating new design options.

## INTRODUCTION

Computational design and simulation tools have helped architects and engineers to improve building performance in recent decades. Parametric generative systems are capable of producing multiple design alternatives; however, to find the best design solutions, optimization tools need to consider different and mostly conflicting design objectives through the alternation of design variables (Kumar *et al.* 2017). Multi-objective optimization methods face complicated and time-consuming challenges to reach the desired results. On one hand, the fitness functions need to converge multiple quantifiable and sometimes conflicting objectives, and on the other hand, many subjective non-quantifiable goals may be overlooked (Yan et al. 2015). The process of optimization requires step by step calculations and consequently demands for significant computation time. Although essential improvements are carried out in terms of optimization and building assessment tools, designers still need to follow a "forward" procedure of defining energy-oriented design parameters and their value ranges (Rezaee et al. 2018; Shahsavari et al. 2019). The optimization methods also spend much time to explore a large number of options and experiments, known as design space, to find the desired solutions; however, in many cases, the architects urge for specific performance objectives though uncertain of the design parameters and their effective value ranges. A generative system may have the potential to explore and learn the design vocabulary through analyzing the provided dataset of existing solutions without the "explicit supervision of the designer" (Bidgoli et al. 2018).

The advent of Artificial Intelligence (AI) technique has brought a new realm of data analysis and applications in different fields of science and engineering. The idea of machine learning is concerned with training computer programs capable of automatic improvement through experience (Mitchell 1997). Above that, Deep Learning (DL) is a "representation learning method" which allows computational models to learn variables much deeper through multiple hidden layers that extract the hidden characteristics of variables known as latent variables (Lecun et al. 2015). Successful applications of deep learning in image recognition, object detection, autonomous vehicles, drug discovery, and image reconstruction provide a promising context for its use in architectural design and optimization.

## BACKGROUND

Architects and engineers apply different methods to find the desired/optimized design solutions. In this section each method is briefly reviewed.

**Optimization tools**

In order to achieve optimized design performance solutions, optimization algorithms need to engage with design alternatives, building energy simulation tools, and quantitative assessment techniques (Kheiri 2018). Based on the review done by Evins (2013) most of the recent studies use meta-heuristic algorithmsto achieve optimized design solutions. There are several building optimization tools using meta-heuristic algorithms, including but not limited to: GenOpt (Wetter 2001), GENE_ARCH (Caldas 2008), MOBO (Palonen et al. 2013), MultiOpt (Pernodet et al. 2011), ParaGen (Turrin et al. 2011), Galapagos (Mcneel 2013), Octopus (Vierlinger 2014), and Optimo (Rahmani et al. 2015).

Optimization tools still face many challenges in converging multiple conflicting objectives that need step by step performance simulations and consequently yield in a noticeable amount of computation time. On the other hand, the optimization tools could only find solutions within the fixed search space that has been defined by designers through parametric modeling methods.

**Machine Learning (ML):**

The considerable computation time used by optimization methods has brought many studies' attention to the application of ML in predicting building performance instead of conventional time-consuming simulations. A few research projects are done using ML algorithms in architectural design and optimization. In this section some existing projects are briefly reviewed.

Zemella et al. (2011) adopt an evolutionary approach using a Neural Network (NN) to discover the optimum design solutions of a typical office building envelope module to reduce the building's energy consumption. The study tests the performance of the developed algorithms in achieving an energy efficient outcome for both single objective and multi-objective optimizations (Zemella et al. 2011). In a study done by Harding et al. (2011), two ML algorithms are applied to the process of designing an exhibition layout that can house flexible exhibition configurations. The study employs a self-organizing map (SOM) to arrange different exhibits and convert them into a spatial plans. An unsupervised neural network is used to classify the design options to different clusters based on their spatial topologies (Harding et al. 2011). Rahmani et al. (2016) create a Form-based Energy Performance Regression Model (FEPRM) framework, to provide an energy performance feedback to the user based on the building's form transformation in the early stages of design. In his study, three ML algorithms are employed for a regression model framework, and the outputs are compared (Rahmani et al. 2016).

The studies demonstrate that ML application has brought development in building design optimization and has reduced computation time significantly; however, in many cases, design alternatives contain a plenty of complex latent variables that may not be analyzed in simple ML algorithms with a shallow network structure. Deeper networks may perform better in analyzing the latent space. Moreover, most of the existing ML methods are made for predicting performance results or clustering dataset, but not generating design solutions. The architects still follow a forward technique in generating the desired/optimized solutions; in contrast, a generative system may have the potential to learn and explore the design syntax and vocabulary through analyzing existing datasets and come up with novel options that may not necessarily exist in the primary input dataset.

**Generative algorithms using Deep Learning (DL) methods:**

ML algorithms with a shallow structure are incapable of learning complicated functions with a high level of abstractions (Bengio 2009). Most dataset with a large number of complex parameters could be investigated through Deep Neural Network (DNN) models that contain multiple layers of latent variables to be applied successfully in different domains including object recognition, information retrieval, classification and regression tasks (Salakhutdinov 2015). Furthermore, based on DL methods, deep generative models such as Pixel Recurrent Neural Network (RNN), autoencoder (AE), variational AE (VAE), and different types of Generative Adversarial Networks (GAN) have brought researchers' attention to generating or reconstructing samples based on the existing dataset. GAN has particularly shown a promising development in recent years. In a study done by Yang et al. (2017), new melody compositions are generated based on specific melody rhythms as training dataset using a novel GAN model (Yang et al. 2017). In another study, Elgammal et al. (2017) have developed a creative GAN capable of generating novel artistic paintings from established styles (Elgammal et al. 2017).

Since the advent of generative models and their promising results, a few studies have conducted practical research in applications of deep learning generative models in design and architecture (Imdat et al. 2018; Bidgoli et al.2019; Rahbar et al, 2019; Chaillou 2019; Newton 2019).

The studies are still in progress and may not show concrete results; however, the promising results of using DL algorithms in other experiments such as image processing, face detection, novel music composition, and artistic paintings generation, to name a few, motivate the architects to study DL as a tool to improve and probably transform the design thinking process. Moreover, none of the above-mentioned studies use DL for design optimization, which is the goal of our study.

## METHODS

The research methods consist of literature review (partially presented in the last section), creating prototypes using DL algorithms, and experiments with the prototypes' applications in prediction and generation of images and design forms, and analyses of the experiment results. The implementation of the prototypes consists of applying the following tools: the Anaconda Python programming platform and Keras Deep Learning packages in the backend, and parametric modeling tools, Rhino/Grasshopper in the frontend.

**Experiment 1: Shape Classification using Convolutional Neural Network (CNN) and synthesized imaging data:**

CNN is a deep neural network algorithm that is commonly applied to image or object classification. CNN is highly efficient in analyzing visual imagery compared to its peers' multilayer perceptrons such as the fully connected Neural Network (Lecun and Bengio 1995). The convolutional layers scan images with multiple filters to analyze the images' specifications such as edges, corners, shapes, etc., which result in various feature maps that constitute those images. Figure 1 shows the structure of a CNN used for this experiment.

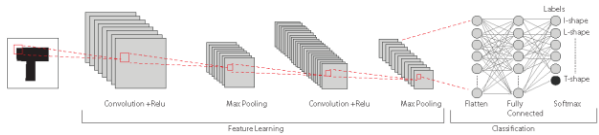

*Figure 1  CNN structure based on Krizhevsky et al. (2012)*

Experiment 1 is done in 3 steps:

1. Generating 6000 images of simple 2D shapes (which could potentially represent architectural plans or any design form outlines), with 6 shape class labels, automatically using Rhino/Grasshopper's parametric modeling capability with random variations.

2. Training the CNN model for the synthetic shapes and their labels as the input dataset (accuracy: 96 % after 10 training epochs).

3. Testing the trained model for 45 manually drawn shapes in Photoshop (accuracy: 93%).

Figure 2 shows the major steps from left to right and top to bottom:

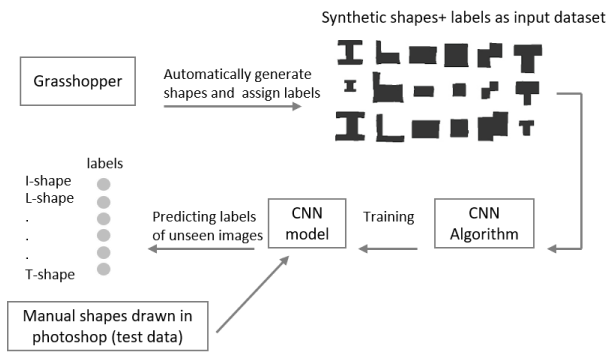

*Figure 2 Predicting the labels of drawn shapes in Photoshop with the trained CNN model*

The algorithm is tested for our synthetic shape dataset. The shape dataset used for this experiment are images of 100 by 100 pixels of simple, 2D shapes which could potentially represent architectural plan outlines (based on floor plan clustering, e.g. Rodrigues et al. 2017). 6 types of shape are used including I-shape, L-shape, Rectangle, Square, Z-shape, and T-shape. Grasshopper components are implemented to parametrically create random points in a defined domain of a region where the result could be recognized as the demanded shape subjectively. Figure 3 graphically represents the generating process for the square shape with the random points controlled in defined domains. The corner (red) points move randomly in a circle area with a specific radius amount, and the middle (green) points nudge up/down and left/right within their defined domains. The points' locations parametrically change in random variations and yield in different square shapes.

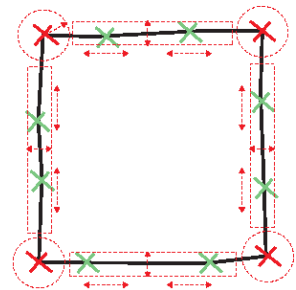

*Figure 3 A sample of synthesized shapes from predefined points which nudge in the controlled areas randomly through parametric modeling*

The shapes are generated in different scales in order to induce more variety and complexity to the dataset. The total number of 6000 images are created and labeled based on their shape types as the input dataset for training. The assumption is that the algorithm analyzes and detects the specifications of the training set and will be able to predict the labels of unseen samples.

In the next step, 45 new shapes are manually drawn in Photoshop with the same image size and color ranges to test the model in predicting the labels of unseen shapes. Figure 4 shows some of the manually drawn shapes in Photoshop.

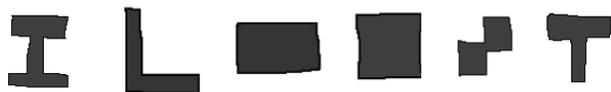

*Figure 4 Sample shapes manually drawn in Photoshop*

**Results and Analysis of Experiment 1:**

The CNN used for this study consists of two convolutional layers with 30 and 15 filters of 5x5 and 3x3 pixels accordingly. The model is trained for 10 epochs with 20 batches. The input dataset is split into training and validation set with a proportion of 3 to 1. The validation set is a representative of the testing dataset to monitor the training process. In Neural Network, a loss function is used during the training to estimate the difference between the predicted and the estimated probability. This error summation, or model's loss value, is calculated in the algorithm based on the difference between the objects' true class labels and the predicted labels and demonstrates how accurate the model works. In a good model, the accuracy improves during the training process while the error is minimized. Figure 5 shows the model's accuracy and loss of the training and validation dataset during the training process.

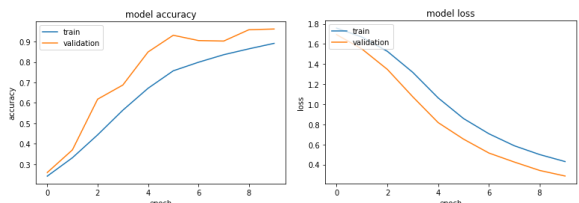

*Figure 5 Left: Accuracy-epoch curve for training and validation datasets; Right: Loss-epoch curve for training and validation datasets.*

The results show that the validation set reaches to good accuracy after 10 training epochs (96%). The validation loss also arrives to a good value of 0.28 after the training process. The experiment shows good result during the training process. It also shows satisfying result during the test process of predicting the labels of new images drawn in Photoshop (accuracy: 93%).

**Experiment 2: Generating shapes with the desired labels using Auxiliary Classifier GAN (AC-GAN):**

In the next experiment, a reverse method is applied to generating shapes with desired labels. AC-GAN is adopted in this study. AC-GAN is a type of GAN algorithm that can not only generate new specific data as GAN, but also can generate the data with the demanded labels (Odena, and Olah, 2016). The algorithm is adopted to the shape dataset for generating shapes with specific labels. The AC-GAN algorithm has two networks called a generator and a discriminator that compete with each other for generating specific data. The idea is rooted in the "game theory" where each network attempts to deceive the other one and hence will be trained in this manner (Goodfellow et al. 2014). The generator initially generates images with random noises, and the discriminator tries to discriminate the generated images based on their validity. Through training the model with a sufficient number of epochs, the generator is able to generate such good images that the discriminator cannot detect as invalid. Figure 6 shows the structure of AC-GAN used for this experiment. In this figure, the shapes are representative of the input dataset that is Parametrically Synthesized in Grasshopper (PSG). On the other hand, the generator generates images through Deep Learning, named as Deep Learning Options (DLO).

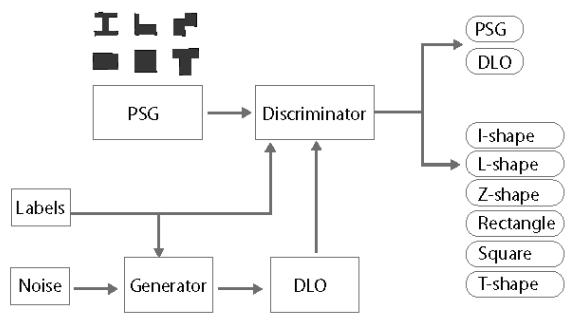

*Figure 6 AC-GAN structure adopted from Mirza and Osindero (2014) and, Odena, Olah, and Shlens (2016)*

The goal is to generate a desired new shape within the limitation of 6 shape class labels. Figure 7 shows the workflow and the generated images by the trained model.

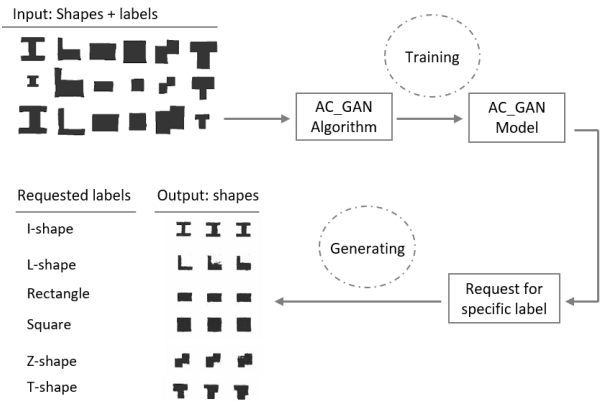

*Figure 7 left to right/top to bottom: Experiment 2 workflow and some examples of the generated dataset using the AC-GAN algorithm, plotted after 4800 training epochs*

**Results and Analysis of Experiment 2:**

The AC-GAN algorithm used in this study consists of two CNN algorithms for generator and discriminator. The model is trained in 5000 epochs with a batch size of 32 for 2616 shapes of the 6 types. When the model is trained, it only takes a few seconds to load the trained model and less than a second to generate the desired shapes. Figure 8 shows discriminator and generator loss over 5000 epochs.

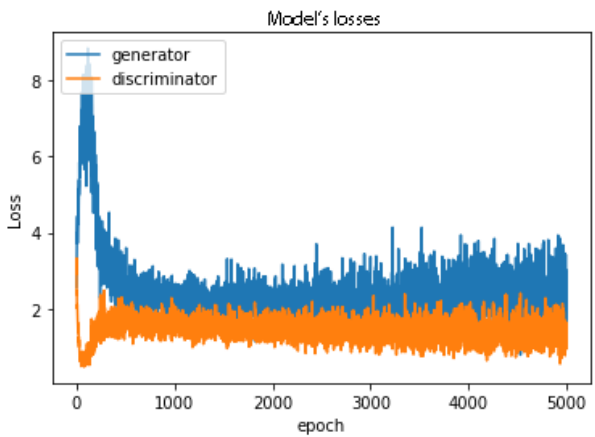

*Figure 8 Loss-epoch curve for the discriminator and generator of the AC-GAN model trained for shape generation*

The discriminator's loss arrives to value 1.1, and the generator's loss reaches to value 2.2 after training process. The trained model shows an overall good performance in generating new shapes of defined labels. Figure 9 shows some shapes generated by the AC-GAN model. In this figure none of the generated shapes originally existed in the input dataset. Specifically, the T-shape shows a novel result that changed the alignment of the left and right edges below its hat. This result presents an innovation in the generator model, because it did not exist in the search space of the original parametric model of the shapes.

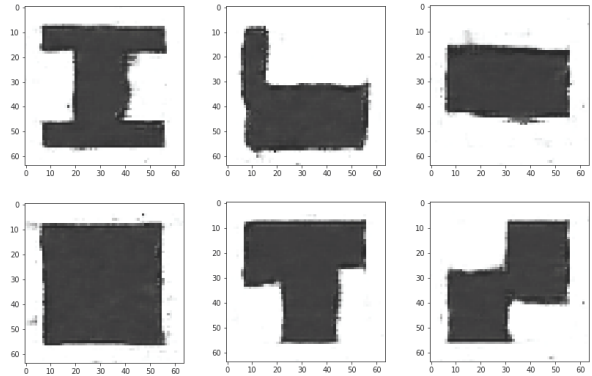

*Figure 9 Some samples of the shapes generated by the AC-GAN model after being trained for 5000 epochs*

**Experiment 3: Generating light/shadow patterns of window/wall based on daylight performance using AC-GAN:**

In this experiment, the same form generation method is applied to generate light/shadow patterns of a hypothetical building facade based on a simple room of 10m x 10m x 4m in Grasshopper. The south facade is divided into 18 by 8 grid of 0.5m × 0.5m cells with a 0.5m margin from each side. The facade has total number of 144 cells which represent a parametric light/shadow pattern of wall/window components. A Python script is used to parametrically light up a random cell in sequence from 1 to 143 in 143 runs accordingly with a repeatable random seed (Run #0 creates 1 window cell and 143 opaque cells, Run #1 creates 2 window cells and 142 opaque cells, …, and Run #142 shows 143 window cells and 1 opaque cell). An annual daylight simulation is done with DIVA4 which provides Spatial Daylight Autonomy (sDA 300, 50%) of the plane work for each pattern parametrically. sDA demonstrates the percentage of the work plane, that for more than 50% of occupied hours receives 300 lux or more for a defined period of time (Illuminating Energy Society, IES). In this study, sDA is only used for labeling the day-lighting performance. Glare metric is not studied for this experiment. Based on the IES standard, the sensor plane is located 0.75 m above the ground with 0.6 m sensor spacing.

Figure 10 top shows some examples of the synthetic patterns of light/shadow used as the training dataset and,

Figure 10 bottom shows the digital model drawn in Grasshopper after daylight simulation for one synthesized pattern.

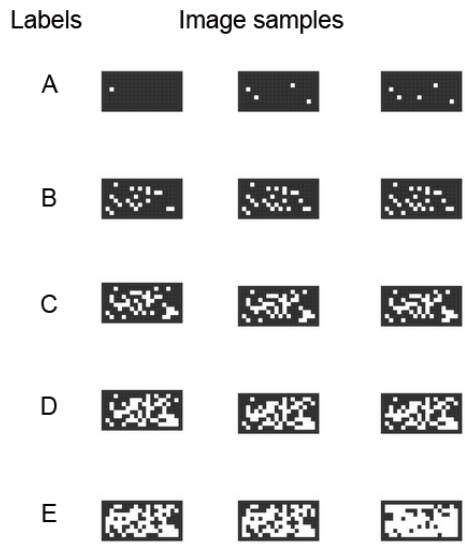

*Figure 10 Top: synthetic pattern images of light/shadow used as input dataset;*

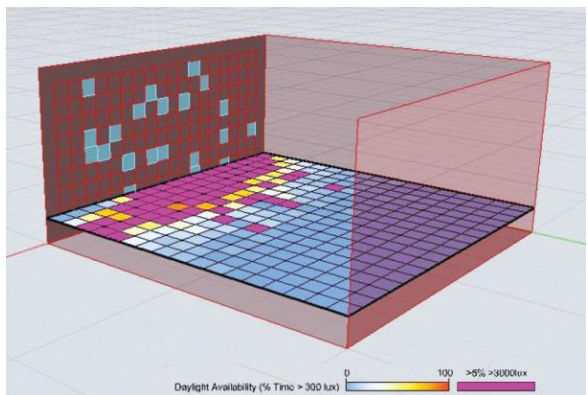

*Figure 10 Bottom: digital model showing one generated pattern simulated with Diva4*

All simulations are done for annual sDA using common building materials with the Houston, Texas TMY weather file. Every simulation is done parametrically with each window/wall pattern generated in Grasshopper, and the sDA metric for each pattern is automatically saved after each simulation. All sDA results are then saved as labels in a text file after completion of each set of runs where each run consists of 143 runs of window/wall pattern with a specific random seed. The whole experiment is repeated for 4 sets of runs with different random seeds ( seed #0 to seed #3) to generate various random patterns as the input dataset.

The total number of 572 (4 × 143) patterns are synthesized with daylight simulation performance. Later a new labeling is automatically mapped to each pattern based on the following category (e.g. sDA of 20% means only 20% of the floor area receives 300 lux or more for more than 50% of the occupied hours annually):
- 0% ≤ sDA < 20%, label = 'A'
- 20 % ≤ sDA < 40%, label = 'B'
- 40% ≤ sDA < 60%, label = 'C'
- 60% ≤ sDA < 80%, label = 'D'
- 80% ≤ sDA ≤ 100%, label = 'E'

The patterns with mapped labeling are given to the AC-GAN algorithm as input dataset. The assumption is that the trained DL generative model can generate new images of façade pattern (window/wall) based on demanded label i.e. daylight performance.

**Results and Analysis of Experiment 3:**

The model is trained in 12,000 epochs with a batch size of 5 for 572 patterns of light and shadow. Figure 11 shows discriminator and generator losses over 12,000 epochs.

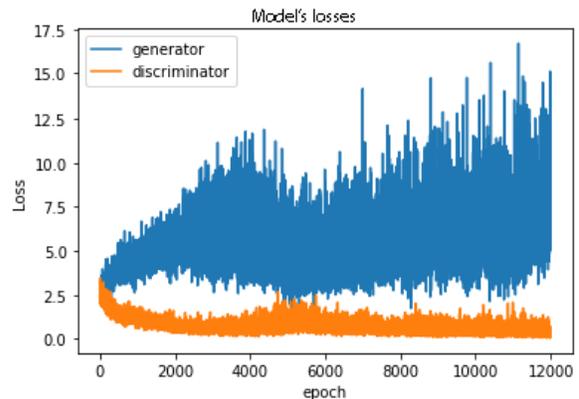

*Figure 11 Loss-epoch curves for the discriminator and generator of the AC-GAN model trained for generating the light/shadow pattern*

The loss value of the discriminator model arrives to 0.2 and the loss value of the generator model reaches 5.0 after the training process. The increase of generator's loss means that the generated patterns are discriminated as fake by the discriminator model during the training process. Based on the labeling method applied to the input dataset, the WWR increase form label "A" to label "E". The generator loss shows the results are not as desired, however, from examination of the results, we can see the match of the window-to-wall ratios (WWRs, from small to large) with the daylight sDA performance labels (A to E). Figure 12 top shows the original images generated by the trained model and Figure 12 bottom shows the same patterns after post processing in which pixels are rounded to absolute black and white.

From the results it can also be found that the generated patterns are not within the search space of the original parametric model of the patterns, therefore can be regarded as novel design options that meet the demanded performance. This experiment shows less variety in generating patterns for each label, for example, the model generates nearly similar patterns for the last two labels (D and E).

Table 1 shows the comparison of daylight sDA performance of Parametrically Synthesized in Grasshopper (PSG) patterns (seed#0 to seed#4) with one generated sample through DL model based on WWR for each label.

Through the comparison in Table 1 between the DL/GAN outcomes (4$^{th}$ and 5$^{th}$ columns) and the PSG results (2$^{nd}$ and 3$^{rd}$ columns), we can see the correlations between WWR and sDA are the same: when WWR increases, sDA performance improves, and vice versa. However, the windows' locations in the façade patterns also affect the sDA performance (and currently the locations were generated randomly for both), therefore we shouldn't expect that the DL/GAN result's WWR to fall necessarily in the range of the PSG results' WWR, for each of the label.

To evaluate the daylight performance of the images generated with DL model, the post-processed images are re-drawn in Grasshopper using the same grid of 0.5m × 0.5m. Erosion and Dilation methods are applied to eliminate the small blobs while keeping the same proportion of black to white. Figure 13 shows a sample pair of images with the same black to white ratio. The left image is generated with DL model after being post-processed and the right one is the corresponding image re-drawn in Grasshopper.

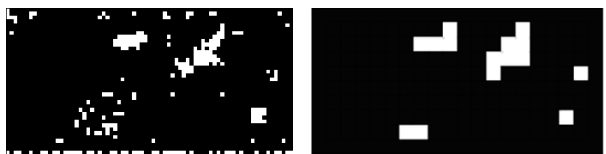

*Figure 13 Left: Image generated with DL model; Right: Image re-drawn in Grasshopper*

The 5$^{th}$ column in Table 1 shows the performance of the re-drawn images based on SDA metrics using DIVA4 daylighting simulation. However, because the location of window and opaque matters in daylight simulation, the re-drawn strategy may not necessarily represent the precise sDA of the DL generated images. Considering the 1st column shows the true labels of each category, the recent claim is demonstrated through the label evaluation of the DL-generated images based on WWR (column 6th), and sDA performance (column 7th) separately. The predicted labels based on WWR and sDA show that the results of the system working for the labels A and E, but having errors for B, C, and D. Note that this is based on the comparison of one sample for each label. In the future work, we will generate multiple samples and re-evaluate the results using the Confusion Matrix method (Stehman 1997) to get a more accurate evaluation of the system.

Moreover, it is worth to mention that the image dataset in this experiment is different from the typical experiments for the CNN models (generator and discriminator) to analyze. In this experiment, each image pattern does not represent specific shapes with recognizable features for the CNN filters that usually extract various features such as edge, corners, shapes and geometry of an image. Considering the complication of the dataset, the model did a reasonable job.

## CONCLUSIONS AND FUTURE WORK

The research has presented two DL/GAN methods for generating desired shapes and façade patterns, based on given shape class labels and building performance labels, respectively. Importantly it demonstrates the new method of utilizing synthesized training data through parametric modeling and simulation that architects are getting familiar with. Parametric modeling, 2D shapes, images, 3D geometry, renderings, etc. representing design options and their corresponding building performance measures through simulations can all become big data for training DL/GAN models, with the aim of generating performance-based, yet innovative design solutions, more efficiently than ever before.

Unlike the optimization tools that only lead to resulting designs within a fixed search space, the demonstrated method using GAN is trained through the search space but is able to produce novel design options. For example, in Experiment 2, the trained model could innovatively generate shapes with given labels.

Experiment 3 needs further studies to improve the performance. Multilayer perceptron algorithms such as DNN might work better for the discriminator and generator of this model; however, the current result shows that the trained model could successfully generate novel patterns with WWRs that match given performance labels. It is worth to emphasize the model's capability in producing new options with the demanded specifications but out of the original search space of design options. The results could then be developed by the designers to reach a more compelling design solution that meets aesthetic criteria as well. Other novel methods such as Style-GAN could be applied to combine the performance-based results with artistic or designer-preferred patterns.

In future studies, the authors intend to (1) improve the current DL/GAN-based generative design system using synthesized big data, and (2) integrate Building Information Modeling (BIM), performance simulation, and optimization for investigating DL/GAN-based generative design methods.

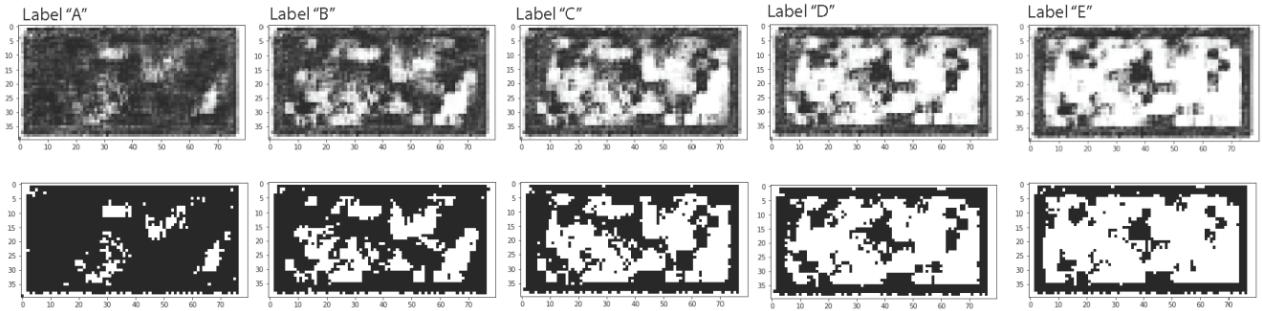

*Figure 12 Top: Some samples of the patterns generated by the trained AC-GAN model; Bottom: The same images after simple post processing (cleaning noises).*

*Table 1 Comparison of daylight sDA performance (Label) between PSG patterns of different seeds and one generated sample through AC-GAN model (shown in Figure 13) based on WWR*

| TRUE LABEL | RANGE VARIANCE OF THE WWR OF PSG PATTERNS FOR SEEDS 0 TO 4 (%) | RANGE OF DAYLIGHT PERFORMANCE OF PSG PATTERNS sDA (%) | WWR OF ONE IMAGE GENERATED BY DL MODEL (%) | DAYLIGHT PERFORMANCE EVALUATION OF THE GENERATED IMAGES WITH DL MODEL sDA (%) | PREDICTED LABEL BASED ON WWR | PREDICTED LABEL BASED ON SDA |
|---|---|---|---|---|---|---|
| A | 0.5 - 11 | 0 - 20 | 7.5 | 12.8 | A | A |
| B | 9 - 21.5 | 20 - 40 | 25 | 53 | C | C |
| C | 17.5 - 30.5 | 40 - 60 | 45 | 72.7 | E | D |
| D | 29 - 40.5 | 60 - 80 | 57 | 100 | E | E |
| E | 38.5 - 71.5 | 80 - 100 | 61 | 100 | E | E |